\documentclass[journal]{vgtc}                

\ifpdf
  \pdfoutput=1\relax                   
  \usepackage{graphicx}                
  \DeclareGraphicsExtensions{.pdf,.png,.jpg,.jpeg} 
\else
  \usepackage{graphicx}                
  \DeclareGraphicsExtensions{.eps}     
\fi%

\graphicspath{{figures/}{pictures/}{images/}{./}} 

\usepackage{subfigmat}

\usepackage{microtype}                 
\PassOptionsToPackage{warn}{textcomp}  
\usepackage{textcomp}                  
\usepackage{mathptmx}                  
\usepackage{times}                     
\usepackage{cite}                      
\usepackage{tabu}                      
\usepackage{booktabs}                  
\usepackage{color}
\usepackage{appendix}
\usepackage{bm}

\newcommand{\ra}[1]{\renewcommand{\arraystretch}{#1}}
\newcommand{\revise}[1]{\textcolor{black}{#1}}

\onlineid{1117}

\vgtccategory{Research}
\vgtcpapertype{algorithm/technique}

\title{Directionally Decomposing Structured Light for Projector Calibration}

\author{Masatoki Sugimoto, Daisuke Iwai, \textit{Member, IEEE}, Koki Ishida, \\ Parinya Punpongsanon, \textit{Member, IEEE}, and Kosuke Sato, \textit{Member, IEEE}}
\authorfooter{
\item
 Masatoki Sugimoto, Daisuke Iwai, Koki Ishida, Parinya Punpongsanon, and Kosuke Sato are with the Graduate School of Engineering Science, Osaka University. E-mail: see https://www.sens.sys.es.osaka-u.ac.jp/.
\item
 Daisuke Iwai is with PRESTO, Japan Science and Technology Agency.
}

\shortauthortitle{Sugimoto \MakeLowercase{\textit{et al.}}:Directionally Decomposing Structured Light for Projector Calibration}

\abstract{Intrinsic projector calibration is essential in projection mapping (PM) applications, especially in dynamic PM.
However, due to the shallow depth-of-field (DOF) of a projector, more work is needed to ensure accurate calibration.
We aim to estimate the intrinsic parameters of a projector while avoiding the limitation of shallow DOF.
As the core of our technique, we present a practical calibration device that requires a minimal working volume directly in front of the projector lens regardless of the projector's focusing distance and aperture size.
The device consists of a flat-bed scanner and pinhole-array masks.
For calibration, a projector projects a series of structured light patterns in the device.
The pinholes directionally decompose the structured light, and only the projected rays that pass through the pinholes hit the scanner plane.
For each pinhole, we extract a ray passing through the optical center of the projector.
Consequently, we regard the projector as a pinhole projector that projects the extracted rays only, and we calibrate the projector by applying the standard camera calibration technique, which assumes a pinhole camera model.
Using a proof-of-concept prototype, we demonstrate that our technique can calibrate projectors with different focusing distances and aperture sizes at the same accuracy as a conventional method.
Finally, we confirm that our technique can provide intrinsic parameters accurate enough for a dynamic PM application, even when a projector is placed too far from a projection target for a conventional method to calibrate the projector using a fiducial object of reasonable size.
} 


\CCScatlist{ 
 \CCScat{K.6.1}{Management of Computing and Information Systems}%
{Project and People Management}{Life Cycle};
 \CCScat{K.7.m}{The Computing Profession}{Miscellaneous}{Ethics}
}

\teaser{
  \centering
  \includegraphics[width=\linewidth]{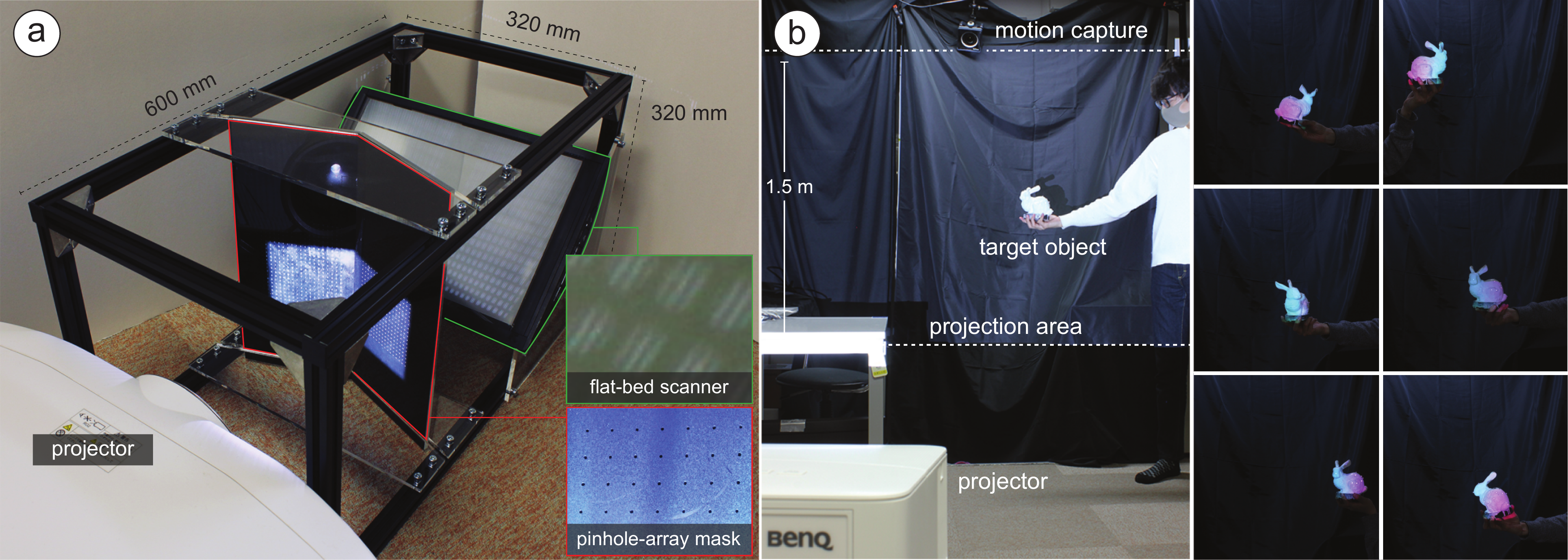}
  \caption{We aim to calibrate a projector's intrinsic matrix while avoiding the shallow depth-of-field limitation of projectors. (a) A prototype calibration device consisting of two pinhole-array masks and a flat-bed scanner can calibrate a projector of any aperture size and focusing distance when placed next to the projector lens. The device directionally decomposes structured light and allows us to regard the projector as a pinhole projector. (b) Our technique accurately calibrated a projector in a dynamic projection mapping application where a color ramp texture was mapped onto a Stanford bunny. This setup requires an impractically large fiducial object to cover the whole projection area when using a conventional calibration technique.}
  \label{fig:teaser}
}

\vgtcinsertpkg

\begin{document}


\firstsection{Introduction}
\label{sec:introduction}

\maketitle

Projection mapping, or spatial augmented reality (SAR), seamlessly merges real and virtual worlds by superimposing computer-generated graphics onto physical surfaces~\cite{Bimber:2005:SAR,doi:10.1111/cgf.13387}.
It has been applied in various fields, such as medicine~\cite{00000658-201806000-00024}, industrial design~\cite{8797923}, online conferencing~\cite{8172039}, office work~\cite{Iwai2011,10.1145/1959826.1959828}, and entertainment~\cite{6193074}.
Owing to recent advances in high-speed and low-latency projector hardware~\cite{watanabe2015high}, the latest research trends toward dynamic projection mapping (PM), where a projection target and/or a projector move under projection~\cite{7516689,10.1145/3272127.3275045,doi:10.1111/cgf.13128,sueishi16}.
To align a projected image onto a target in a dynamic environment, the projection image must be geometrically corrected.
According to the perspective projection principle, geometric correction requires the shape of a projection surface, the pose of the surface relative to a projector (extrinsic matrix), and the projector's intrinsic matrix. 
We can accurately estimate the shape using a range scanner.
Although a projector itself cannot measure the state of a target, an external sensor such as a motion capture system can estimate the extrinsic matrix.
However, more work is needed to accurately calibrate the intrinsic parameters.

\begin{figure}[t]
  \centering
  \includegraphics[width=0.98\linewidth]{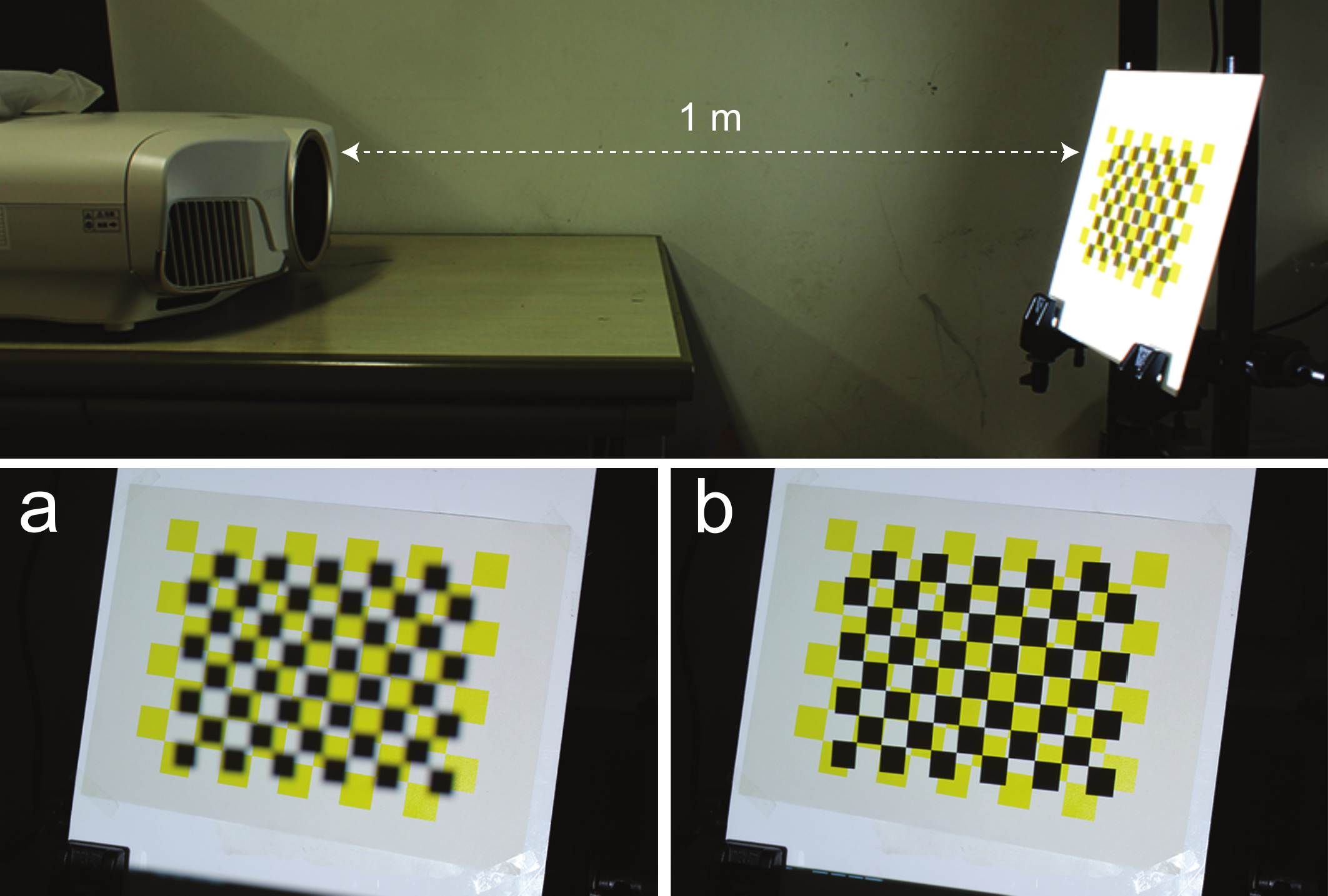}
  \caption{The projected spatial pattern is defocused on a handy checker board placed 1 m from a projector when the focusing distance is 3 m (a). It appears focused when the focusing distance is 1 m (b).}
  \label{fig:eg_defocus}
\end{figure}

The primary barrier to calibrating intrinsic parameters is the shallow depth-of-field (DOF) of a projector.
Projectors are generally designed with large apertures to maximize their brightness, which on the other hand, leads to significantly shallow DOFs.
Calibration requires dense correspondences between three-dimensional (3D) world coordinates and two-dimensional (2D) projector image coordinates.
The correspondences are obtained by projecting spatial light patterns onto a fiducial object such as a checkerboard and measuring the projected patterns with an external sensor such as a camera.
In principle, intrinsic parameters can be accurately calibrated when the correspondences are obtained over the whole projector image coordinate.
Due to the shallow DOF of a projector, the fiducial object needs to be placed close by the projector's focusing distance, where a target surface is assumed to be placed in each application.
When the fiducial object is placed even a slightly different from the focusing position, the projected spatial light patterns appear significantly defocused (Fig.~\ref{fig:eg_defocus}).
This does not cause a serious problem in calibration for an application whose working volume is near the projector because a handy fiducial object can cover the whole projector image coordinate.
However, as shown in a previous study~\cite{7781767}, an impractically large fiducial object is required to calibrate a projector in an application, in which a user moves a target surface at a large distance from the projector (e.g., $\geq$3 meters).
This limitation has prevented the expansion of the application field in dynamic PM.

We propose a calibration technique that estimates the intrinsic parameters of a projector while avoiding the limitation of shallow DOF.
As the core of our technique, we present a practical calibration device that requires a minimal working volume directly in front of the projector lens regardless of the projector's focusing distance and DOF.
The device consists of a flat-bed scanner and pinhole-array masks (Fig.~\ref{fig:teaser}a).
The technique calibrates a projector by projecting a series of structured light patterns such as gray codes in the device.
The pinholes directionally decompose the structured light, and only the projected rays that pass through the pinholes hit the scanner plane.
For each pinhole, we extract a ray (known as a chief ray in the field of optics) passing through the optical center of the projector.
Consequently, we can regard the projector as a pinhole projector that projects only the extracted chief rays.
Thus, we calibrate the projector from the extracted rays based on the standard camera calibration technique introduced by Zhang et al.~\cite{888718} that assumed a pinhole camera model.
The paper presents the design of our calibration device as well as the entire calibration process, including the chief ray extraction technique.
Using a proof-of-concept prototype, we demonstrate that our technique can calibrate projectors with different focusing distances at the same accuracy as Zhang's technique.
Finally, we validate the practical advantages of our technique compared with a conventional calibration procedure through a large-scale dynamic PM experiment, in which Zhang's technique cannot be directly applied due to the necessity of an impractically large checkerboard.

In summary, our primary contributions are that we
\begin{itemize}
    \item Introduce an intrinsic projector calibration technique that overcomes the shallow DOF limitation by extracting the chief rays from defocused projection light,
    \item Develop a practical calibration device that directionally decomposes projected structured light using pinhole-array masks,
    \item Extract chief rays from the decomposed light, by which the calibrated projector can be regarded as a pinhole projector, and
    \item Implement a proof-of-concept prototype that demonstrates accurate intrinsic calibrations of projectors without requiring impractically large fiducial objects.
\end{itemize}

\section{Related Work}

The camera and the projector have the same perspective projection model, owing to their optical duality between them~\cite{10.1145/1073204.1073257}.
The standard camera calibration procedure using a pinhole camera model~\cite{888718} has thus far been applied to projector calibrations~\cite{5204319,6375029,10.1117/1.2336196,9107397}.
Because a projector cannot sense the real world, a camera was used in earlier methods to measure projected spatial patterns on a fiducial object.
However, due to the shallow DOF of a projector, these methods require impractically large fiducial objects in large-scale PM applications.
When projectors and projection targets are static, self-calibration or auto-calibration techniques can calibrate the extrinsic and intrinsic parameters of the projectors by projecting structured light patterns onto the target surfaces without requiring fiducial objects~\cite{5710903,5981781,6060818,8115403}.
However, these techniques cannot be used in dynamic PM applications, in which, generally, projection surfaces do not initially exist.

A major solution to the shallow DOF limitation is a coaxial setup of a projector and a camera using a beam splitter~\cite{7516689,10.1145/3272127.3275045,doi:10.1111/cgf.13128,sueishi16,6595980,7831400,IWAI2010162,Punpongsanon2015}.
In a coaxial system, the pixel correspondences between the projector and camera are not changed according to the distance of a projection surface from the lenses.
Thus, once the camera is calibrated, the projector's calibration becomes unnecessary.
However, the beam splitter halves the luminance of projected light.
In addition, a precise alignment of the projector and camera relative to the beam splitter is required~\cite{6910020}.
The devices easily become misaligned due to the vibration or heat caused by extended use.
Thus, the coaxial setup is not preferable in practical applications.

The shallow DOF limitation in projector calibration significantly constrains the working space of dynamic PM systems that do not apply coaxial setups~\cite{8007312,10.1145/2816795.2818111}.
A previous study overcame the limitation and realized a projector calibration procedure in which a user needs only a small calibration board, regardless of the focusing distance of the projector~\cite{7781767}.
The calibration requires correspondences between 2D projector coordinates and 3D world coordinates.
The previous method projects random dot patterns onto the calibration board, and the center of each dot was used to obtain the correspondences.
However, as the earlier study noted~\cite{7781767}, when the dot is defocused, its center is not equal to the actual center of the dot.
In addition, the defocusing error increases with a larger projector lens.
Thus, the previous method potentially suffers from the shallow DOF limitation in calibrating for a large-scale PM, which generally requires a bright projector consisting of a bright light source and a large aperture.
On the other hand, our method works in principle in such situations (i.e., calibration errors do not increase with a large aperture.)
Another previous study applied a line stripe pattern projection to obtain the correspondences~\cite{Li:14}.
Projected line stripes are defocused on a calibration board placed near the projector lens.
The center of each line can be obtained as the pixels having local peak luminances, which are then used in a phase-shifting algorithm to obtain the correspondences.
Unfortunately, this technique works only in situations where the projected line stripes are properly defocused so that the center of each line can be correctly detected.

Our method is inspired by a previous camera calibration technique that applied a calibration device consisting of pinhole-array masks and a liquid-crystal display (LCD) panel~\cite{8099504}.
The previous method displayed gray code patterns on the LCD and captured emitted rays through the pinholes.
The camera's aperture was minimized so that only the chief rays hit the camera's image plane.
Finally, the LCD coordinates of the chief rays were obtained by decoding the gray code and used to calibrate the camera based on Zhang's method~\cite{888718}.
This calibration procedure is similar to our method.
However, although obtaining the chief rays can be easily achieved in the previous study, doing so is more difficult in our case because an aperture is normally not adjustable in commercially-available projectors.
As one of our study's primary contributions, we show how to extract the chief rays of a projector from all the projected rays passing through the pinholes and hitting the scanner plane of a calibration device.

\section{Method}

Our method is based on Zhang's standard camera calibration method assuming a pinhole-camera model~\cite{888718}.
\revise{In Appendix~\ref{appendix}, we explain how to extend the camera calibration technique to projector calibration.}
Our technique applies the computational part of the projector version of Zhang's method.
In this section, we introduce the principle of directionally decomposing structured light using pinhole-array masks.
We then describe the core of our technique: extracting chief ray from the decomposed information.
Finally, we explain how to estimate the intrinsic parameters of a projector using the extracted chief rays.


\subsection{Directionally decomposing structured light using pinhole-array masks}
\label{subsec:decomp}

\begin{figure}[t]
  \centering
  \includegraphics[width=0.98\hsize]{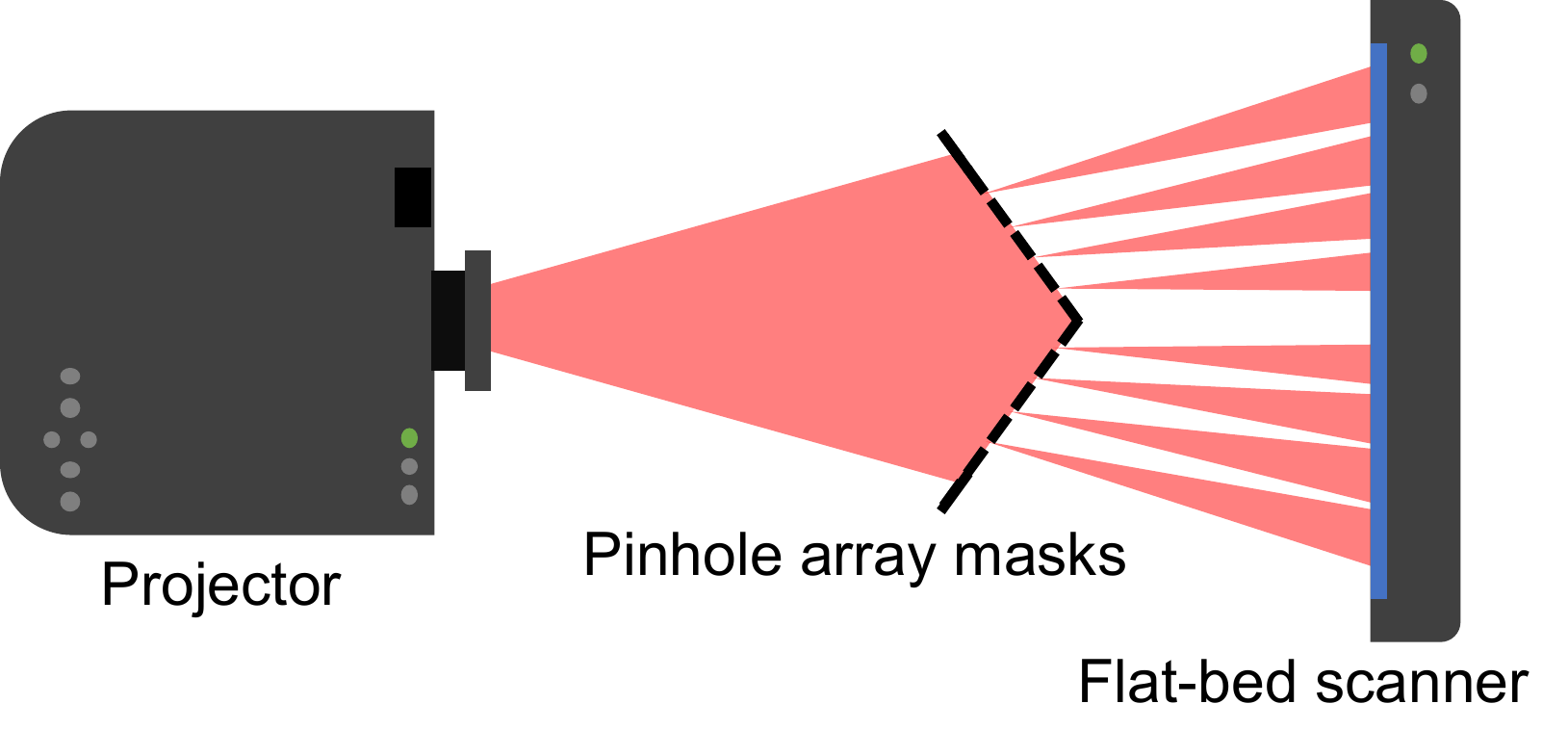}
  \caption{Configuration of our calibration device and an example of projected light rays in the device (red).}
  \label{fig:system_concept}
\end{figure}

Our calibration device consists of planar pinhole-array masks and a flat-bed scanner.
We place the masks to meet the following two requirements: (1) they are located between a projector to be calibrated and the scanner, and (2) they do not occlude each other when viewed from the projector's point of view (Fig. \ref{fig:system_concept}).
Because parallel planes do not provide additional constraints for Zhang's method, the masks are rotated by several degrees from the scanner plane and the other masks.
In detailing our method, we assume that the number of the pinhole-array masks is two, which preserves generality.

Zhang's method accurately calibrates a projector as long as the projected calibration patterns are focused on fiducial planes.
Our technique is based on Zhang's method; however, we use the pinhole-array masks and the scanner surface, rather than a checkerboard, as the fiducial planes.
We place the calibration device directly in front of the projector lens.
Consequently, projected calibration patterns are strongly blurred on the pinhole-array masks in most cases (Fig.~\ref{fig:teaser}a). Nevertheless, our technique can always obtain chief rays by directionally decomposing projected light.
Our technique virtually converts a projector of arbitrary aperture sizes and focusing distances to a pinhole projector, by which we can accurately obtain the 2D coordinates of projected pixels corresponding to predefined points on the fiducial planes, such as pinholes, even when projected images are defocused.

\begin{figure}[t]
  \centering
  \includegraphics[width=0.98\hsize]{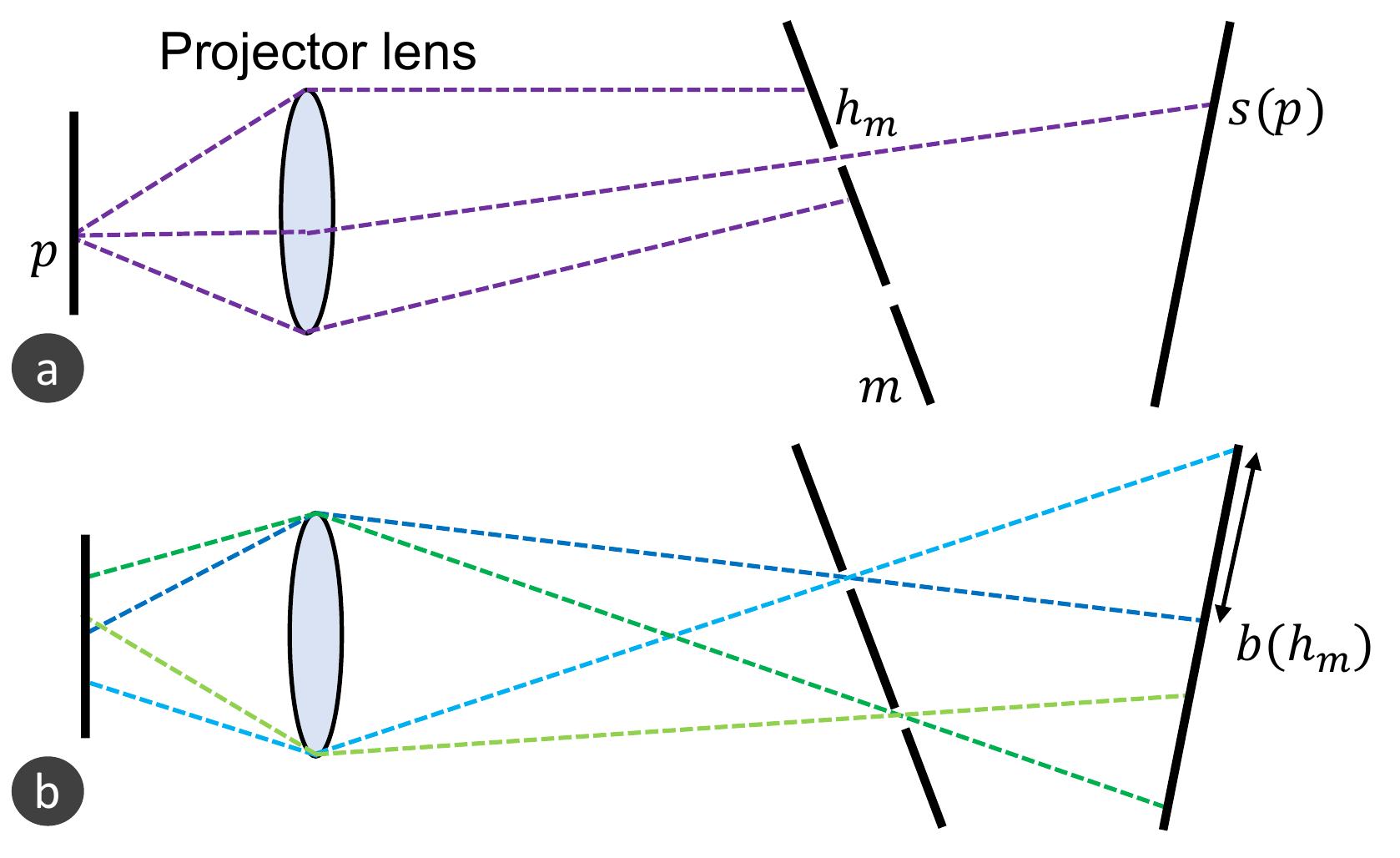}
  \caption{Principle of directional decomposition of projected light using a pinhole-array mask. Different colors of the dashed lines represent rays from different projector pixels. (a) A light ray passes through a hole $h_m$ from a single pixel $p$ and hits a surface point $s(p)$. (b) A light blob $b(h_m)$ appears on the scanner surface consisting of light rays emitted from adjacent projector pixels and passing through the same pinhole $h_m$.}
  \label{fig:principle}
\end{figure}

Our calibration device directionally decomposes projected light as follows. 
As a simple case, suppose a single projector pixel $p$ is turned on (white), and the other pixels are turned off (black).
Light rays emitted from the white pixel are refracted at different parts of the projector lens (Fig. \ref{fig:principle}a).
These refracted rays are then projected onto a pinhole-array mask $m$, where they do not converge into a point, forming a defocused pixel.
When a pinhole $h_m$ locates inside the defocused pixel, one of the rays passes through it and hits a point $s(p)$ on the scanner surface.
In this manner, our calibration device can extract a ray emitted from a projector pixel in a specific direction.

\begin{figure}[t]
  \centering
  \includegraphics[width=0.98\hsize]{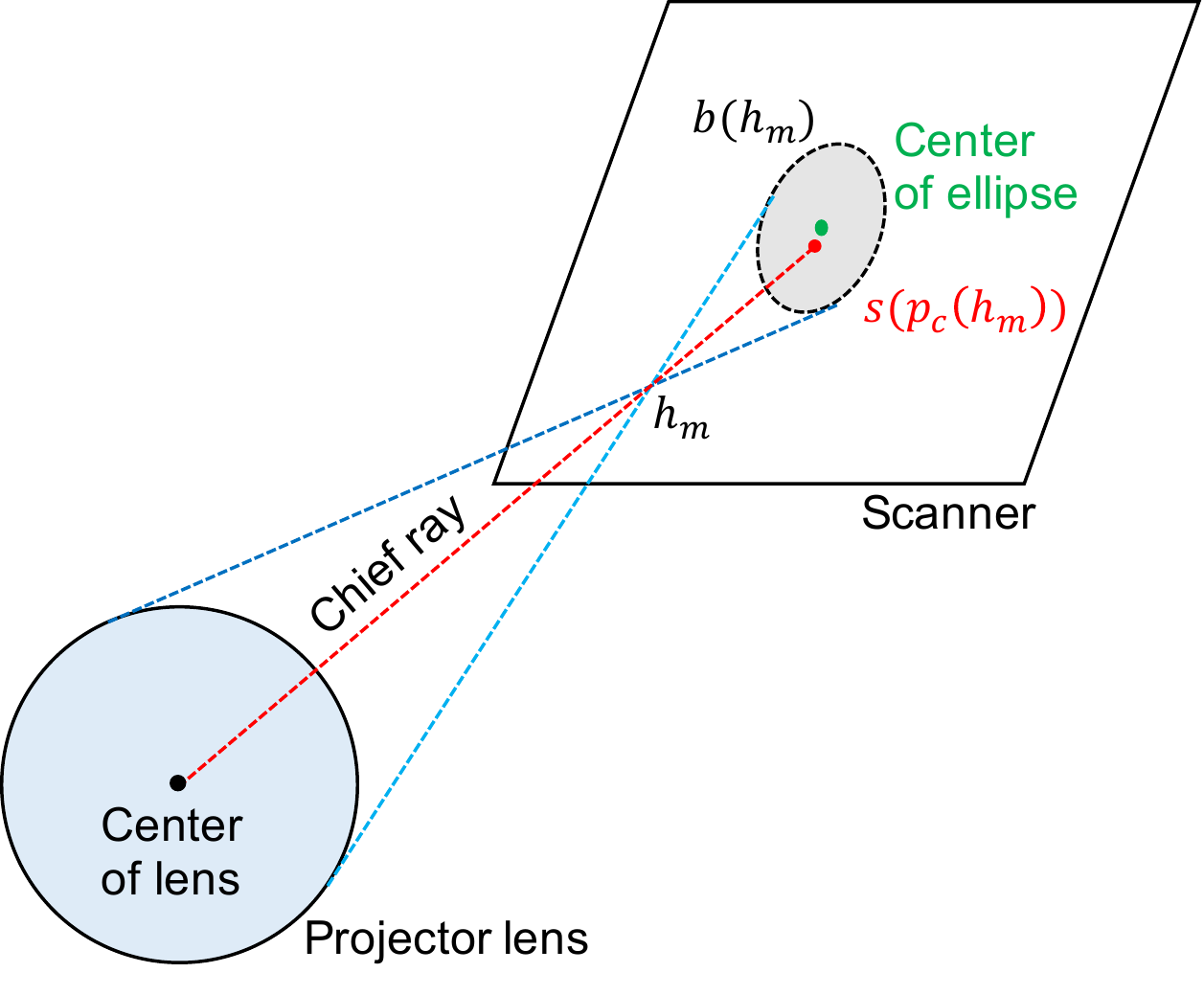}
  \caption{A light blob forms an ellipse, which is the intersection of the scanner surface with an oblique cone whose apex and base are the pinhole $h_m$ and the lens, respectively. The center of the ellipse is not identical to the intersection of the scanner surface with a chief ray.}
  \label{fig:cone}
\end{figure}

When we turn on all the pixels, light blobs appear on the scanner surface.
Each blob is formed by rays that are emitted from adjacent projector pixels and pass through the same pinhole (Fig. \ref{fig:principle}b).
We denote the blob of a pinhole $h_m$ as $b(h_m)$.
The shape of a blob is determined by the shape of the lens aperture and the pose of the scanner relative to the projector.
The rays incident on the edge of a blob come from the edge of the projector lens.
Because a projector's lens aperture is generally a circle, the blob shape is an ellipse whose eccentricity becomes zero if the scanner surface is parallel to the principal plane of the lens.
A blob $b(h_m)$ is the intersection of the scanner surface with an oblique cone whose apex and bases are the pinhole $h_m$ and the lens, respectively (Fig. \ref{fig:cone}).


\subsection{Chief ray extraction}\label{subsect:method_chief}

When using Zhang's method that assumes a pinhole projector, we need to extract a chief ray that passes through the optical center of the lens from all the rays incident on each blob $b(h_m)$.
In other words, we need to identify the projector pixel $p_c(h_m)$ emitting the chief ray. 
However, doing so is not trivial.
The chief ray is identical to the axis of the oblique cone described in Sect. \ref{subsec:decomp} and also shown in Fig. \ref{fig:cone}.
Thus, on the scanner surface, the point where the chief ray is incident (i.e., $s(p_c(h_m))$) is shifted from the center of the blob's ellipse.
Therefore, simply computing the center of the blob does not provide us with the chief ray pixel.
The distance between these two points increases when the eccentricity of the blob's ellipse increases.
If the scanner surface is completely parallel to the principal plane of the projector lens, the blob becomes a true circle, and these two points coincide.
However, it is difficult to manually realize such geometric alignment because a user does not know where the principal plane exists.

We propose a technique to extract the chief ray of each blob, which works even when the shape of the blob is an ellipse of any eccentricity.
Instead of extracting the chief ray on the scanner surface, our technique identifies the projector pixel emitting the chief ray (i.e., $p_c(h_m)$) on the projector's image plane.
In this technique, we assume that the projector's image plane and the principal plane of the lens are parallel, which is generally achieved in the manufacturing of projectors.

\begin{figure}[t]
  \centering
  \includegraphics[width=0.98\hsize]{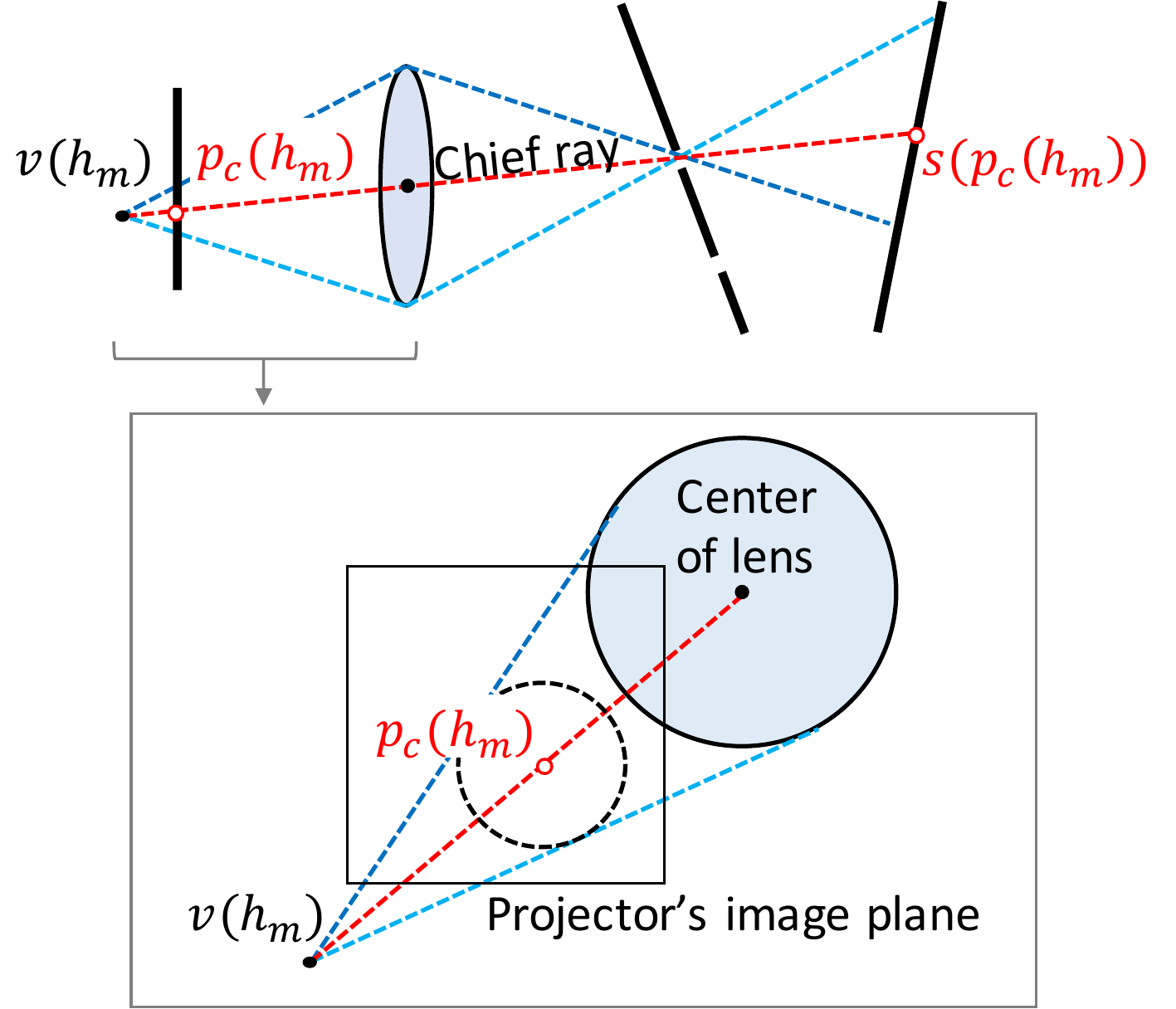}
  \caption{Chief ray extraction by the back-projection of a light blob from the scanner surface onto the projector's image plane.}
  \label{fig:backproj}
\end{figure}

As shown in Fig.~\ref{fig:backproj}, rays incident on a blob $b(h_m)$ can be regarded as emitting from a virtual light source $v(h_m)$ located behind the projector image plane.
We consider an oblique cone whose apex and base are $v(h_m)$ and the projector lens, respectively.
The lateral surface of the cone consists of rays that are incident on and refracted at the edge of the lens.
The rays then pass through the pinhole $h_m$ and finally become incident on the edge of the blob $b(h_m)$.
According to the above-mentioned parallel assumption, the intersection of the cone with the projector image plane forms a true circle.
Therefore, the chief ray is emitted from the projector pixel at the center of the circle.
We identify $p_c(h_m)$ by projecting points in the blob on the scanner surface back to the projector's image plane, fitting a circle to the back-projected blob and computing its center.
The back-projection can be easily done once we determine the 2D coordinates of the projector pixels incident on the blob.
We identify the 2D coordinates by projecting a series of structured light patterns (e.g., gray code patterns), measuring the resultant illuminance patterns on the scanner surface, and decoding the scanned patterns.
Because a back-projected blob forms a circle on the projector image plane regardless of the eccentricity of the ellipse of the blob on the scanner surface, this technique can reliably identify the projector pixel emitting the chief ray in any geometrical relationship between the projector and the scanner.
\revise{Because back-projection is done simply by looking up the correspondences, we do not require any other information, such as the poses of the pinhole-array masks.}



\subsection{Intrinsic parameters estimation}\label{subsec:method_intrest}

\begin{figure}[t]
  \centering
  \includegraphics[width=0.98\linewidth]{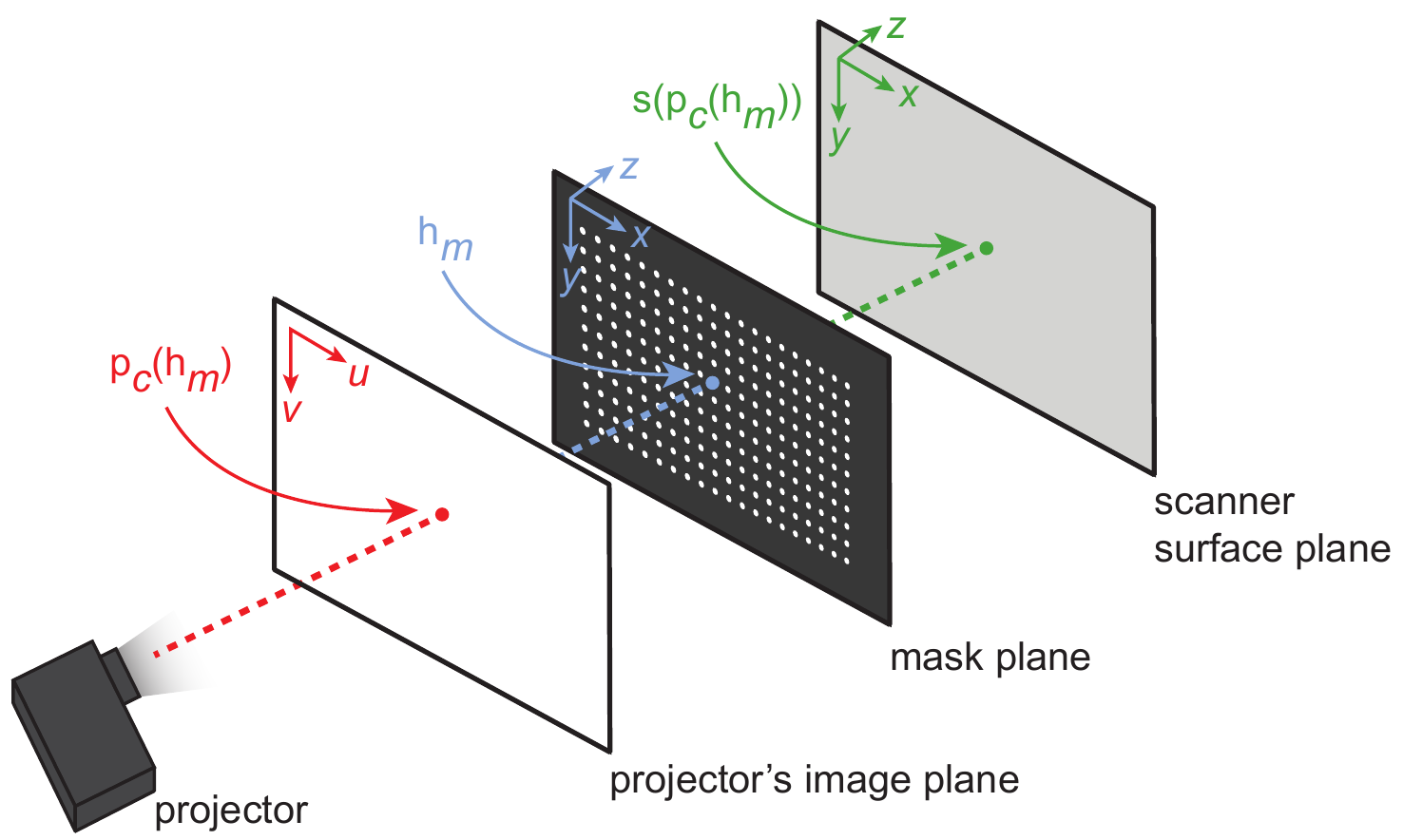}
  \caption{A set of corresponding 3D points on the scanner surface and the mask planes and 2D points on the projector's image plane.}
  \label{fig:corresponding_points}
\end{figure}

Once the projector pixel of the chief ray $p_c(h_m)$ is identified, we obtain the corresponding scanner surface point $s(p_c(h_m))$ by inversely looking up the decoded result of the gray code pattern projection.
We also identify the corresponding pinhole $h_m$ using the following two-step method.
First, we divide the blobs on the scanner surface into two groups, each of which corresponds to each mask $m$.
Because the masks do not overlap from the projector's point of view (Sect.~\ref{subsec:decomp}), the blobs can be divided using a simple clustering algorithm such as K-means.
Second, we apply a circles-grid recognition algorithm to each group to identify the pinholes.
As a result, we obtain a set of corresponding 3D object points on the scanner surface and the mask planes (i.e., $h_m$ and $s(p_c(h_m)$) and 2D image points on the projector's image plane (i.e., $p_c(h_m)$) (Fig.~\ref{fig:corresponding_points}).
We denote each set of corresponding points as $c(h_m)=\{h_m,s(p_c(h_m)),p_c(h_m)\}$.
Sets of corresponding points are then directly used in Zhang's method (see Appendix~\ref{appendix}) to calculate the intrinsic matrix of the projector.

The accuracy of estimated intrinsic parameters potentially decreases primarily due to image noise.
To alleviate this problem, we apply a robust estimation technique and exclude outliers in the calibration parameter estimation.
First, we estimate the intrinsic matrix of the projector and its extrinsic matrix relative to the scanner surface using all the sets of corresponding points.
The intrinsic and extrinsic matrices are denoted as $\tilde{\textbf{K}}_0$ and $[\tilde{\textbf{R}}|\tilde{\textbf{t}}]_0$, respectively.
We then used these matrices to computationally project each projector pixel $p_c(h_m)$ onto the scanner surface, which we denote as $\tilde{s}_0(p_c(h_m))$.
Suppose the reprojection error of each set of corresponding points on the scanner surface is $||s(p_c(h_m))-\tilde{s}_0(p_c(h_m))||$, we denote its mean over all sets of corresponding points as $e_0$.
Then, we exclude a set of corresponding points $c_0(h_m)$, which has the largest reprojection error, and estimate the calibration parameters again, which we denote as $\tilde{\textbf{K}}_1$ and $[\tilde{\textbf{R}}|\tilde{\textbf{t}}]_1$.
After $n$ iterations of this process, we estimate the calibration parameters $\tilde{\textbf{K}}_n$ and $[\tilde{\textbf{R}}|\tilde{\textbf{t}}]_n$ by excluding $n$ sets of corresponding points with the largest reprojection errors (i.e., $c_0(h_m), \ldots, c_{n-1}(h_m)$).
We expect that the averaged reprojection error $e_n$ decreases when $n$ is increased, and, at some point, $e_n$ might start to increase because excluded sets of corresponding points include inliers as well as outliers.
Therefore, when the averaged reprojection error reaches a local minimum, we stop excluding the outliers and regard the estimated intrinsic and extrinsic matrices at this iteration as the most accurate parameters.

\section{Experiment}

In building a prototype calibration device, we validated the proposed calibration method.
This section introduces our prototype and then presents intrinsic calibration results for projectors of different aperture sizes with different focusing distances.
The prototype's calibration accuracy is compared with that of the conventional calibration method.
We then demonstrate how our chief ray extraction method improves the calibration result, and we analyze the outliers.
Finally, we evaluate how the proposed calibration method is accurate for a dynamic PM application.

\subsection{Experimental setup}

\begin{figure}[t]
  \centering
  \includegraphics[width=0.98\linewidth]{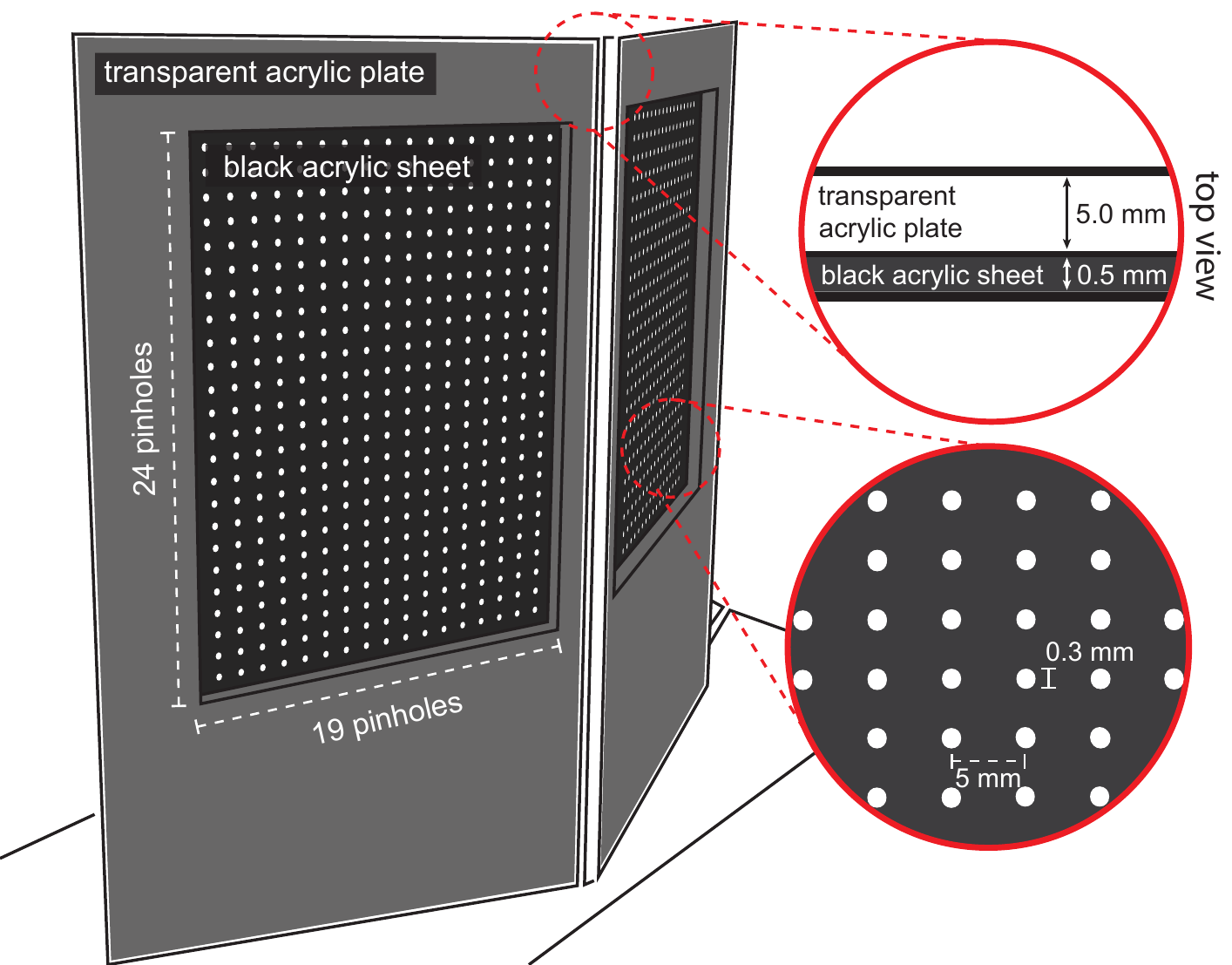}
  \caption{Configuration of the prototype pinhole-array masks.}
  \label{fig:pinhole-assembly}
\end{figure}

We built a calibration device consisting of two pinhole-array masks and a flat-bed scanner (Canon LiDE220, 216 mm$\times$297 mm, 4,800 dpi) (Fig.~\ref{fig:teaser}a).
Each mask consisted of a thin black acrylic sheet (thickness: 0.5 mm) and a thick transparent acrylic plate (thickness: 5.0 mm).
We poked 19$\times$24 pinholes whose diameters were 0.3 mm into the black acrylic sheet at 5 mm intervals using a laser cutter (Universal Laser Systems, ILS9.75).
\revise{The pinhole diameter was experimentally determined so the contrast of the scanned image was not significantly degraded by diffraction.
The inter-pinhole distance was determined as follows.
For accurate calibration, the number of 2D-3D correspondences (i.e., $c(h_m)$) needed to be increased.
This was achieved by increasing the number of pinholes through shortening the inter-pinhole distance. However, a too-short inter-pinhole distance caused an overlap of light blobs on the scanner surface. Considering this trade-off, we experimentally determined the distance of our prototype.}
The transparent acrylic plate was used to support the pinhole sheet.
We made a rectangular hole through the plate using the same laser cutter so that the pinholes were not covered by the transparent acrylic plate, while the black acrylic sheet was rigidly supported when we glued the sheet and the plate together (Fig.~\ref{fig:pinhole-assembly}).
We attached a rear-projection screen film onto the scanner surface so that projected patterns could be sufficiently scattered for measurement by the scanner.
To validate the versatility of the proposed method, we prepared three projectors with different lens diameters and field of views (FOV): (1) a DLP projector (BenQ MS524, 800$\times$600 pixels, lens diameter: 24 mm, FOV: middle), (2) a second DLP projector (NEC NP110J, 800$\times$600 pixels, lens diameter: 33 mm, FOV: wide), and (3) a 3-LCD projector (Epson EH-TW8400, 1920$\times$1080 pixels, lens diameter: 64 mm, FOV: narrow).
In the rest of this paper, we refer to these projectors as (1) DLP1, (2) DLP2, and (3) LCD.
Because we could not directly measure the aperture size of each projector, we manually measured the lens diameter as a substitute for the aperture size.


\begin{figure}[t]
  \centering
  \includegraphics[width=0.98\hsize]{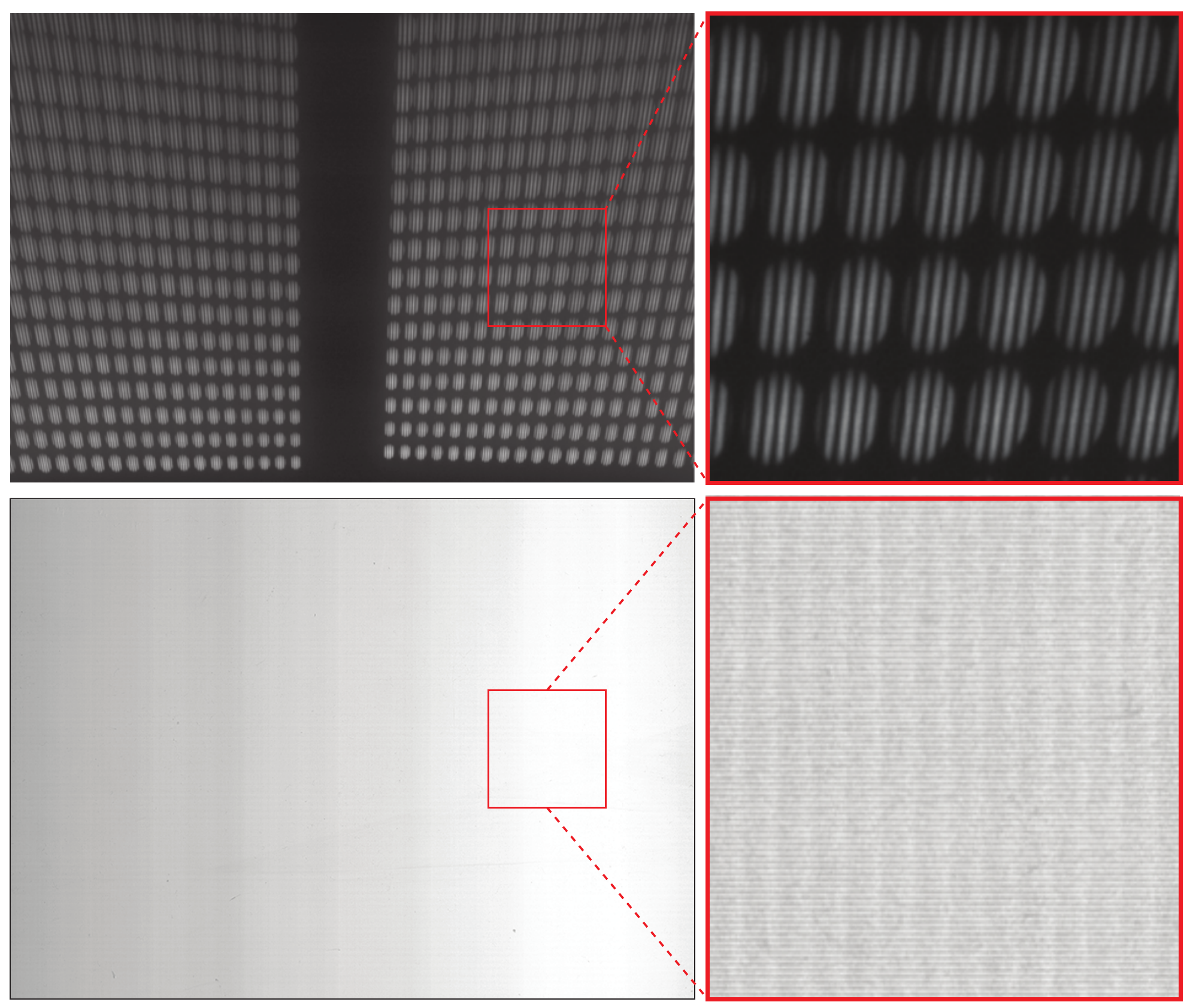}
  \caption{Scanned images of the least significant bit of gray code pattern with the pinhole-array masks (top) and without the masks (bottom).}
  \label{fig:w_wo_mask}
\end{figure}

The masks and the scanner were fixed by black aluminum frames and acrylic jigs, as shown in Fig.~\ref{fig:teaser}a.
All the projectors were placed in the same position.
The distance of a projector from the masks and from the scanner were approximately 250 mm and 450 mm, respectively.
The masks were rotated 20 degrees around the yaw axis, and the scanner was rotated 45 degrees around the pitch axis.
Figure~\ref{fig:teaser}a also shows the appearances of the masks and the scanner surface when the LCD projector projected a fringe pattern.
The pattern was not visible on the masks due to the defocus, while it was visible on the scanner surface and formed light blobs, each of which corresponded to a pinhole.
To confirm the necessity of the pinhole-array masks, we projected the finest fringe pattern (the least significant bit of gray code pattern) and scanned the projected result in two conditions: whether or not the masks were placed between the projector and the scanner.
Figure~\ref{fig:w_wo_mask} shows the scanned results.
When the masks were not placed, the scanned result was almost uniformly white and could not obtain any spatial information of the projected pattern.
On the other hand, a fringe pattern is clearly visible in each blob in the scanned result with the masks.
Therefore, we confirm that the masks effectively decomposed the structured light.

\subsection{Intrinsic calibration}\label{subsec:exp_intrcalib}

We validated our method's calibration accuracy by comparing it with a conventional camera-based calibration approach.
Specifically, the three projectors used in this study were calibrated with the proposed and conventional methods under two conditions regarding the distance from the projectors to the plane of sharp focus.
We adjusted the focusing ring of each projector so that the projected imagery appeared focused 1 m (the near condition) and 3 m (the far condition) from the projector lens.

\begin{figure}[t]
  \centering
  \includegraphics[width=0.98\linewidth]{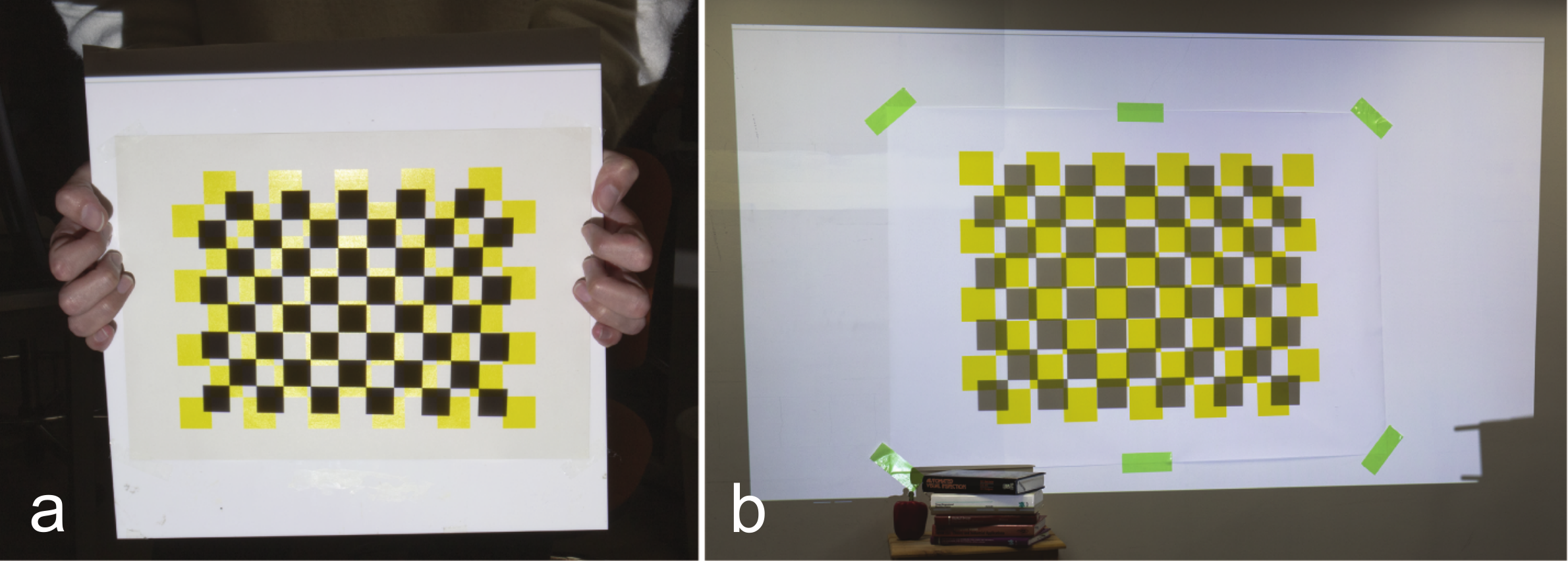}
  \caption{Yellow-and-white checkerboards in the conventional method onto which a black-and-white checker pattern was projected in (a) the near condition and (b) the far condition.}
  \label{fig:conventional_method}
\end{figure}

\subsubsection{Conventional camera-based calibration}

Camera-based calibration was performed as follows.
We printed a yellow-and-white, 7$\times$10 checker pattern on flat sheets of paper whose sizes were A4 and A0 for the near and far conditions, respectively (Fig.~\ref{fig:conventional_method}).
\revise{For accurate calibration, the fiducial object needs to cover the projector’s FOV as much as possible. If the A4 pattern is used for the far condition, it should be placed at many locations to meet this requirement because it is too small to cover the FOV at 3 m.}
The distances of the checker corners were 21.0 and 85.5 mm in the near and far conditions, respectively.
We attached the A4 board onto a flat and rigid acrylic plate, which enabled us to manually change its pose.
On the other hand, the A0 board was attached to a wall in our experimental room because we could not obtain a sufficiently rigid and flat plate for it.
Thus, we could not move it using our hands.
A camera (Canon EOS REBEL T2i, 5184$\times$3456 pixels) was placed next to a calibrated projector.

We performed the calibration of each projector in a dark room using the following steps.
First, we placed the checkerboard and the projector so that the projected imagery appeared focused on the board.
We projected a uniform white image onto the board and captured it with the camera.
We then projected a black-and-white, 7$\times$10 checker pattern and captured the overlaid image.
We performed this process 15 times with different relative poses between the projector and the board, moving the board in the near condition while moving the projector in the far condition.
A printed checker corner (3D object point) was obtained by analyzing the printed pattern of each captured image under white illumination.
A projected checker corner was obtained by analyzing the blue channel of the corresponding captured image under the checker pattern projection.
Using a homography transformation, we computed the 2D projector coordinate value of a pixel projected at the printed checker corner (2D image point).
Finally, using the 2D-3D correspondences, Zhang's method provided the projector's intrinsic parameters.

\begin{table*}[tb]
    \ra{1.1}
	\caption{Estimated parameters and mean reprojection error (MPRE) \revise{in the projector image coordinate system} in the (a) near and (b) far conditions. The unit of the values is pixel of a projector.}
	\label{table:condition1_parameters}
	\centering
	(a)\\
    \begin{tabular}{@{}ccccccccc@{}}
        \toprule
        & \multicolumn{2}{c}{DLP1} & \phantom{a} & \multicolumn{2}{c}{DLP2} & \phantom{a} & \multicolumn{2}{c}{LCD} \\ \cline{2-3}\cline{5-6}\cline{8-9}
        & Proposed & Conventional & & Proposed & Conventional & & Proposed & Conventional \\ 
        \midrule
        $f_x$  &  2047.65    & 2166.65    && 1578.80    & 1505.84 &&  3661.12  & 3512.25 \\
        $f_y$  &  2057.85    & 2149.91  && 1579.76     & 1509.01 && 3671.80    & 3525.51 \\ 
        $c_x$  &  404.29     & 427.01 && 358.25 & 430.59 && 921.89    & 911.99 \\
        $c_y$  &  739.26     & 521.47 && 636.85 & 715.69 && 489.26    & 502.75 \\
        MRPE & 0.59 & 0.30 && 0.73 & 0.41  && 0.59    & 0.37 \\
        \bottomrule
    \end{tabular}
    \\\ \\(b)\\
    \begin{tabular}{@{}ccccccccc@{}}
        \toprule
        & \multicolumn{2}{c}{DLP1} & \phantom{a} & \multicolumn{2}{c}{DLP2} & \phantom{a} & \multicolumn{2}{c}{LCD} \\ \cline{2-3}\cline{5-6}\cline{8-9}
        & Proposed & Conventional & & Proposed & Conventional & & Proposed & Conventional \\ 
        \midrule
        $f_x$   & 2044.18   & 2111.94  && 1599.02    & 1590.90  && 3186.74 & n/a \\
        $f_y$   & 2066.60   & 2151.71  && 1593.01    & 1610.39  && 3222.72 & n/a \\ 
        $c_x$   & 388.21    & 397.89   && 383.97     & 396.79  && 902.36 & n/a  \\
        $c_y$   & 778.79    & 783.47   && 639.39     & 691.18  && 523.67 & n/a  \\
        MRPE  & 0.67 & 0.80 && 0.66 & 0.41  && 0.86 & n/a \\
        \bottomrule
    \end{tabular}
\end{table*}

\subsubsection{Comparing intrinsic parameters and reprojection errors}

Tables~\ref{table:condition1_parameters}a and \ref{table:condition1_parameters}b show the calibration results of the three projectors in the near and far conditions, respectively.
$f_x$ and $f_y$ are the focal lengths of the projector in the horizontal and vertical direction, respectively. $c_x$ and $c_y$ are the coordinates of the projector’s principal point on the image plane.
The MRPE (mean reprojection error) in the tables was computed as the mean of squared distances between the image points and mathematically projected corresponding object points onto the projector's image plane using the calibration results.
In the far condition, we could not calibrate the LCD projector with the conventional method because its size and weight made it impossible to change its pose relative to the checkerboard.
On the other hand, all three projectors could be calibrated in both conditions with the proposed method in which the projectors were placed at the same position in our calibration device in all the cases.

Comparing the calibrated parameters between the proposed and conventional methods, we found that both methods provided similar values, except $c_y$ of the DLP1 and DLP2 in the near condition.
Estimation of $c_y$ in projector calibration is generally unreliable~\cite{5981781}.
Because a projector is manufactured to project images upward in both business and entertainment applications, its optical axis passes below the center of the projected image region or even lower.
Zhang's calibration algorithm requires the initial value of each calibration parameter which should be as close as possible to the ground truth.
The initial value of $c_y$ is generally the center of the image coordinate system, which works well for camera calibration because the optical axis of a camera passes through the center of the camera image region in most cases.
On the other hand, due to the off-axis property of a projector, the same initial value does not provide a stable estimation of the projector's $c_y$.
The LCD projector is unique in that its optical axis passes through almost the center of the projector's imaging plane; thus, the estimated $c_y$ values of the LCD projector are similar in both methods.
Despite difference in $c_y$ estimations, the MRPE values were less than 1 pixel in all the projectors and conditions. 
In summary, we confirm that the proposed method's calibration performance is comparable to the conventional method for different types of projectors with different aperture sizes (lens diameters). 

\begin{figure}[t]
  \centering
  \includegraphics[width=0.95\linewidth]{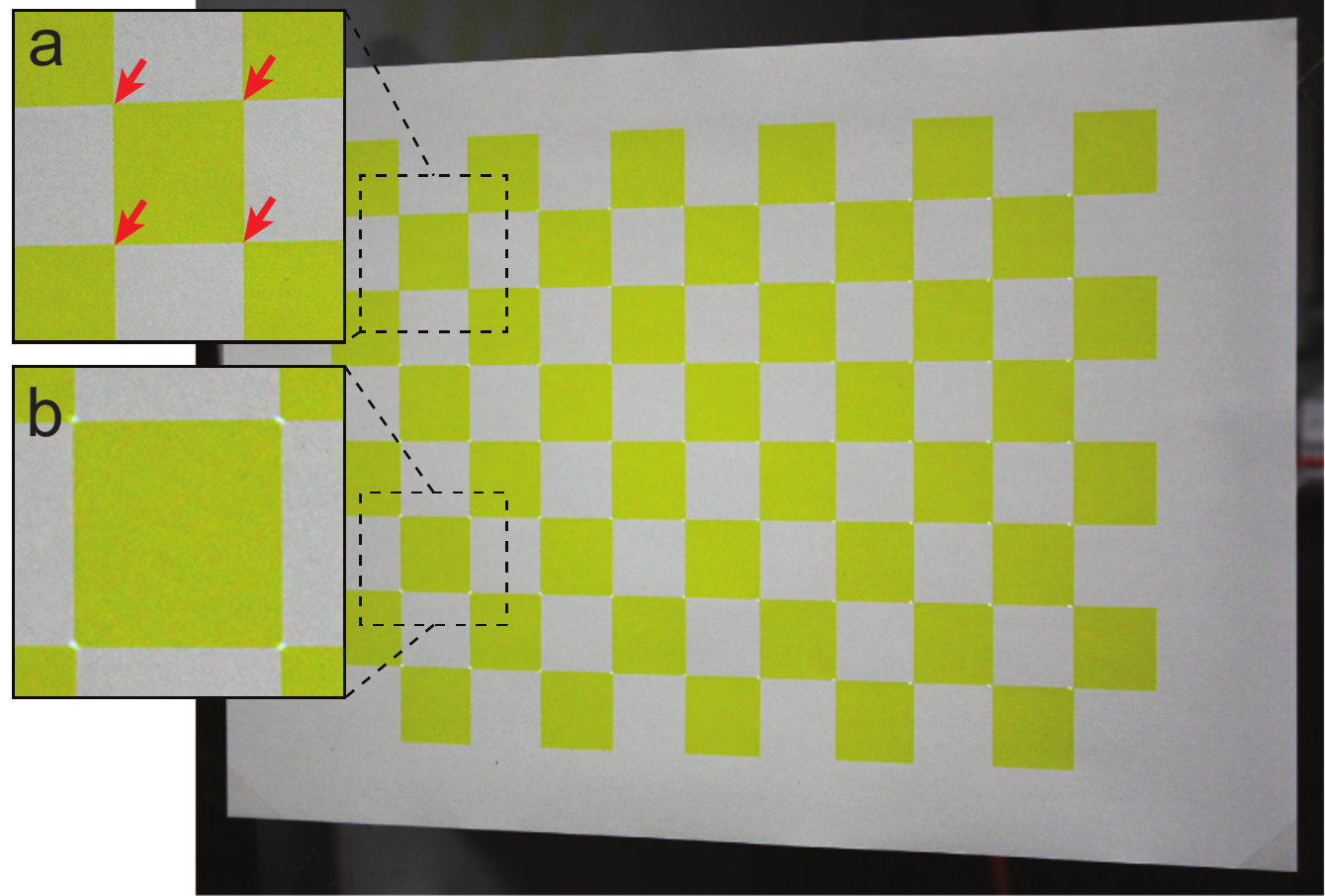}
  \caption{Using a checkerboard to evaluate calibration accuracy assuming dynamic PM scenarios. (a) The checker corners at the top-left area of the board (indicated by red arrows) were used to compute its pose relative to a projector to be evaluated. Using the pose and calibrated intrinsic parameters, projector pixels corresponding to the other checker corners were computed. (b) The distances between the checker corners and the corresponding projected pixels were evaluated.}
  \label{fig:evaluation_board}
\end{figure}

\subsubsection{Calibration accuracy evaluation assuming dynamic PM scenarios}

Because the RPE evaluates the calibration accuracy only on the fiducial planes used in the calibration, a low RPE does not mean that a calibration method is useful in dynamic PM, where a projection target might be placed where the fiducial planes were not placed.
Therefore, we further assessed calibration accuracy by estimating the pose of a calibrated projector relative to a checkerboard placed at an unknown location using the calibrated parameters.
We used the A4 and A0 checkerboards in the near and far conditions, respectively.
First, we projected a series of gray code patterns onto the board and captured the projected results with the camera.
This provided us the projector coordinates of the four checker corners at the top-left section of the board (Fig.~\ref{fig:evaluation_board}a).
Second, we estimated the projector pose relative to the board by solving the Perspective-n-Point (PnP) problem using the four correspondences and the calibrated intrinsic parameters.
Third, we projected dots onto all the checker corners except the four used in solving the PnP.
The dot positions were computed using the estimated pose and the intrinsic parameters.
Finally, we measured the physical distance of each projected dot from the corresponding checker corner on the board (Fig.~\ref{fig:evaluation_board}b).



\begin{figure}[t]
  \centering
  \includegraphics[width=0.98\linewidth]{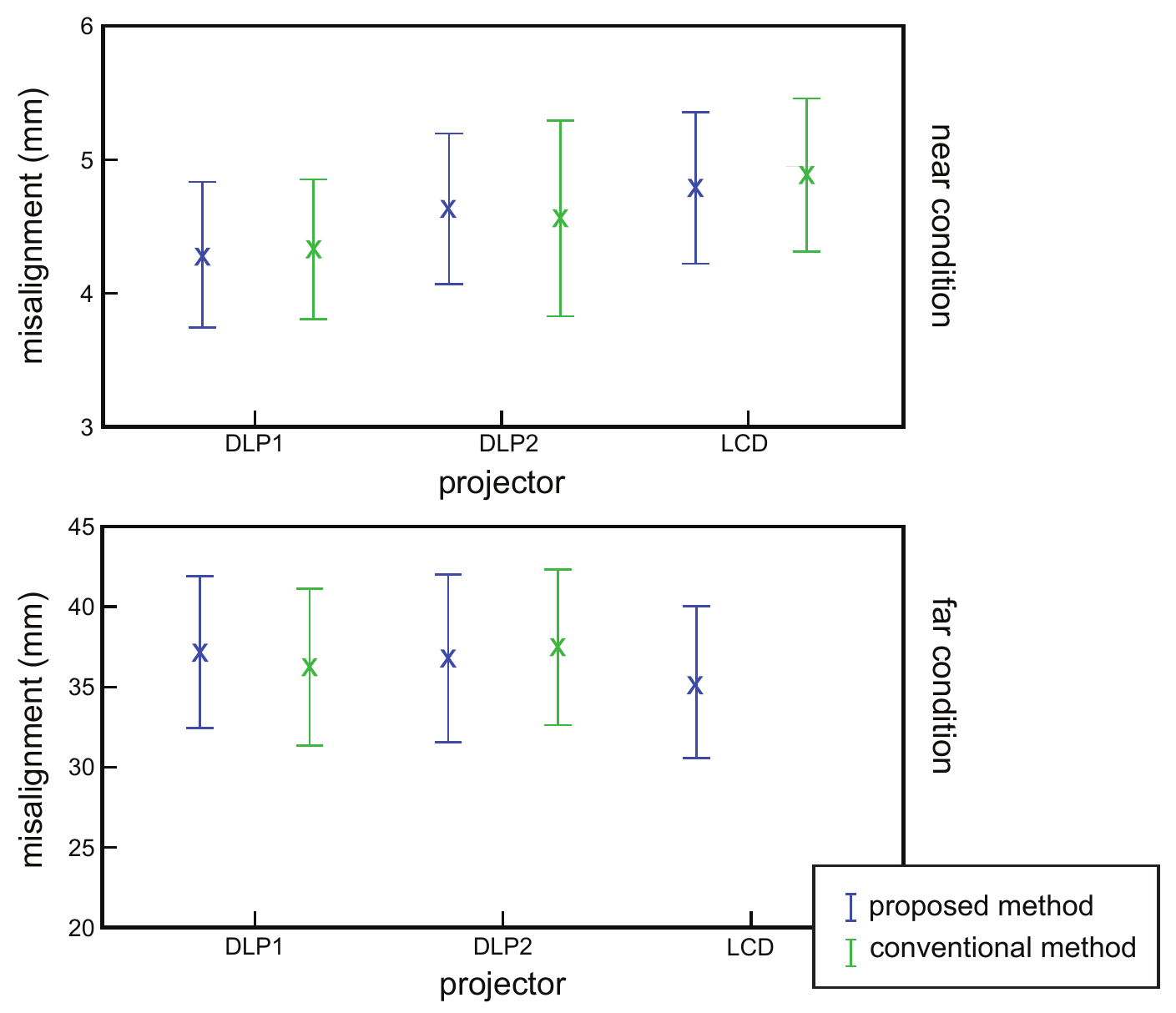}
  \caption{Mean and standard deviation of distances between checker corners and projected dots.}
  \label{fig:projected_error}
\end{figure}

Figure~\ref{fig:projected_error} shows the mean and standard deviation of measured error distances.
Because the LCD projector could not be calibrated using the conventional method in the far condition, the figure presents no data.
We conducted a paired $t$-test to compare the error distances between the proposed and conventional methods for each projector in each condition.
We found no significant difference between the methods in any pairs ($p\geq0.05$).
\revise{Therefore, we find no significant difference in projection accuracy between our method and conventional methods in dynamic PM.}

\subsection{Chief ray extraction}

We validated the proposed chief ray extraction technique.
As described in Sect.~\ref{subsect:method_chief}, a blob on the scanner surface forms an ellipse when the scanner surface is not parallel to the principal plane of the projector lens.
In this case, the chief ray is not incident on the center of the ellipse.
Thus, we extract a projector pixel emitting the chief ray by back-projecting the blob onto the projector's image plane, on which the blob appears as a true circle whose center corresponds to the chief ray pixel.
We compared the calibration results of our method with and without the proposed chief ray extraction technique.

\begin{figure}[t]
  \centering
  \includegraphics[width=0.98\linewidth]{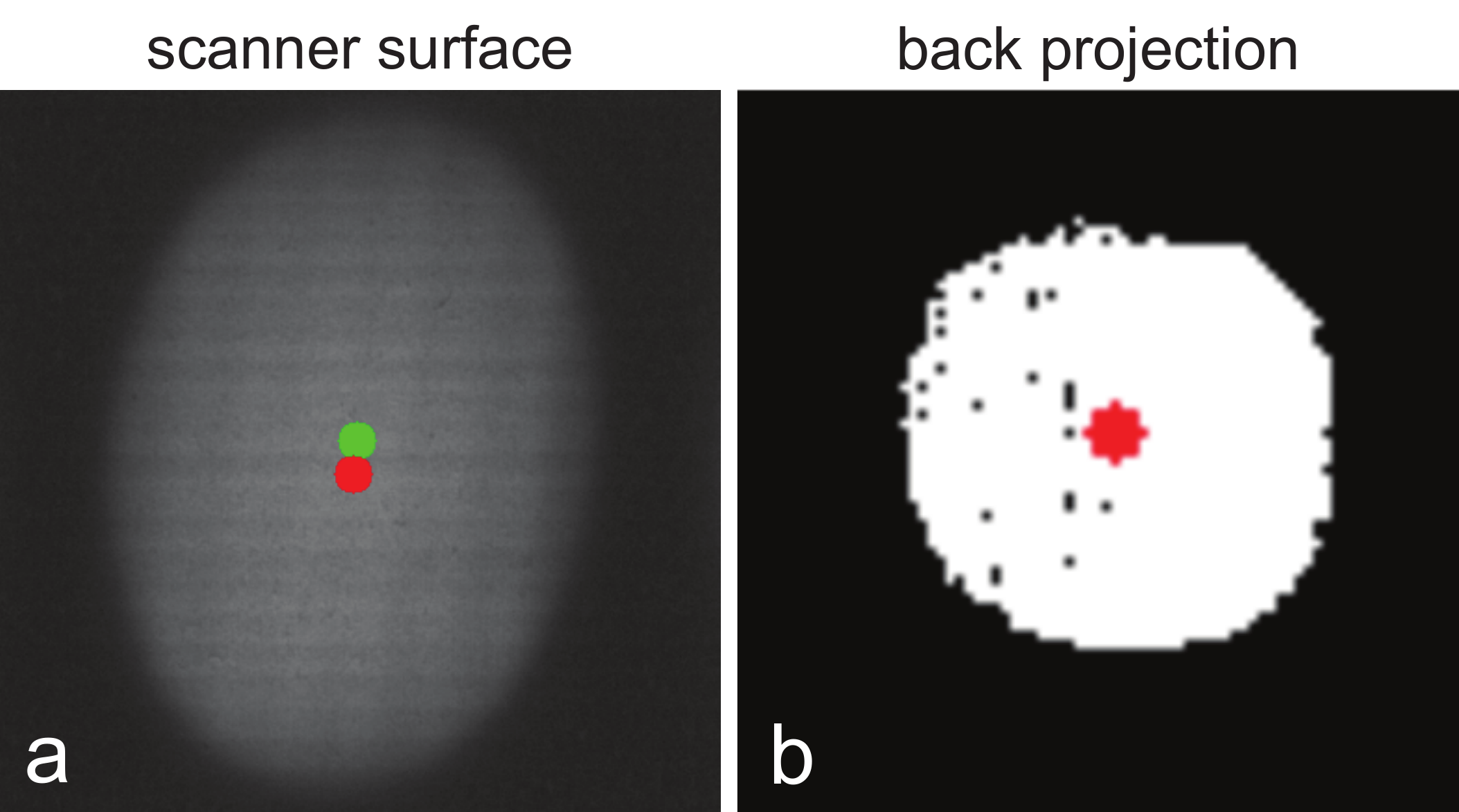}
  \caption{A light blob on (a) the scanner surface and (b) its back-projection on the projector's image plane. The green dot in (a) indicates the center of a fitted ellipse. The red dot in (b) indicates the extracted chief ray pixel, and the red dot in (a) indicates the corresponding point on the scanner surface.}
  \label{fig:res_chief_ray}
\end{figure}

\begin{table}[tb]
    \ra{1.1}
	\caption{Estimated parameters and MPRE \revise{in the projector image coordinate system} of the DLP1 projector with the focusing distance of 1 m. The ``proposed (na\"{i}ve)'' column shows the parameters estimated by regarding the center of each ellipse on the scanner surface as a chief ray pixel. The ``proposed'' and ``conventional'' columns are copied from Table~\ref{table:condition1_parameters}. The unit of the values is pixel of a projector.}
	\label{table:comparison_chiefray}
	\centering
    \begin{tabular}{@{}cccc@{}}
        \toprule
        & proposed (na\"{i}ve) & proposed & conventional \\ 
        \midrule
        $f_x$  & 2054.56 &  2047.65    & 2166.65 \\
        $f_y$  & 2072.16 &  2057.85    & 2149.91 \\ 
        $c_x$  & 375.78 &  404.29     & 427.01 \\
        $c_y$  & 747.28 &  739.26     & 521.47 \\
        MRPE & 1.32 & 0.59 & 0.30 \\
        \bottomrule
    \end{tabular}
\end{table}

Figure~\ref{fig:res_chief_ray}a shows a scanned blob when a uniform white image was projected from the DLP1 projector with a focusing distance of 1 m.
The blob formed an ellipse whose center is indicated by a green dot in the figure.
Figure~\ref{fig:res_chief_ray}b shows the back-projected blob on the projector's image plane, which formed a circle rather than an ellipse.
The center of the circle is indicated by a red dot in the figure.
The red dot is also overlaid on Fig.~\ref{fig:res_chief_ray}a, which indicates the scanner surface point corresponding to the center of the back-projected blob.
As we expected, the green and red dots are not identical in Fig.~\ref{fig:res_chief_ray}a.
The ``proposed (na\"{i}ve)'' column of Table~\ref{table:comparison_chiefray} presents the calibration result of the DLP1 projector in the near condition when we assume that the chief ray hits the green dot.
For comparison, we present the calibration results of the same projector in the near condition from Table~\ref{table:condition1_parameters}a.
We found that the calibration parameters did not vary significantly among the three methods except $c_y$.
As discussed in Sect.~\ref{subsec:exp_intrcalib}, the variance of $c_y$ among the methods is common in projector calibration.
On the other hand, the MRPE was much larger (over 1 pixel) in the proposed method with the na\"{i}ve chief ray extraction than the other methods.
In general, a calibration result with an MRPE of larger than 1 pixel is a poor result and not recommended for use in an application.
Therefore, we confirm that our chief ray extraction technique, based on the back-projection of the scanned blobs onto the projector's image plane, is essential in providing an accurate calibration result.

\subsection{Outlier analysis}

\begin{figure}[t]
    \centering
        \includegraphics[width=0.98\linewidth]{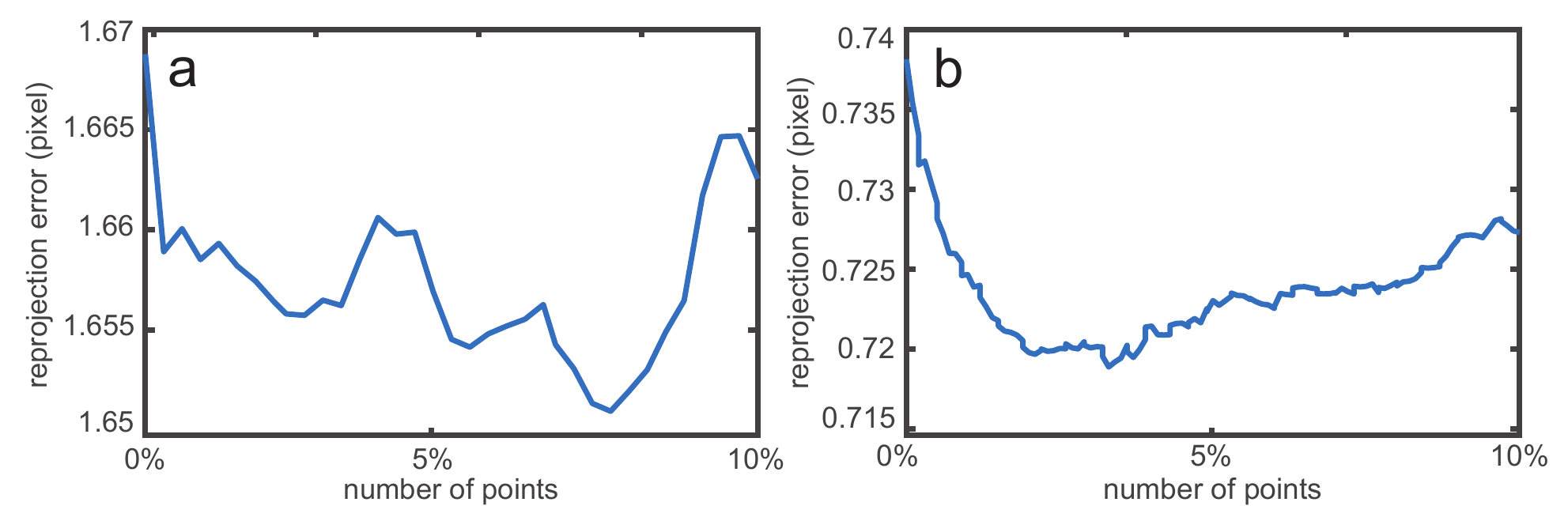}
    \caption{Reprojection errors \revise{in the scanner image coordinate system} at different numbers of excluded outliers in the calibrations of (a) DLP2 and (b) LCD projectors in the near condition. The unit of the values is pixel of the scanner.}
    \label{fig:reprojection-error}
\end{figure}

As described in Sect.~\ref{subsec:method_intrest}, our method applies a robust estimation technique in the intrinsic parameter calibration.
Figure~\ref{fig:reprojection-error} shows the reprojection errors in the scanner image coordinate system at different numbers of excluded outliers in the calibrations of the DLP2 and LCD projectors in the near condition.
The range of the number of outliers was from 0\% to 10\% of all the sets of corresponding points $c(h_m)$.
We excluded the correspondence in order from large to small reprojection error.
As expected in Sect.~\ref{subsec:method_intrest}, when increasing the number of outliers, the errors initially decreased before beginning to increase.
Specifically, the error began to increase after 7.6\% and 3.3\% of all the sets of corresponding points were excluded in the DLP2 and LCD data, respectively.
The same trends were confirmed in the results of the other projectors and the other conditions.
Therefore, we confirm that the outlier exclusion positively affected our calibration method.
The number of outliers to be excluded should be carefully determined.
Specifically, we recommend using the local minimum of the reprojection errors by increasing the number of outliers up to 10\% of all the sets of corresponding points.

\begin{figure}[t]
  \centering
  \includegraphics[width=0.98\linewidth]{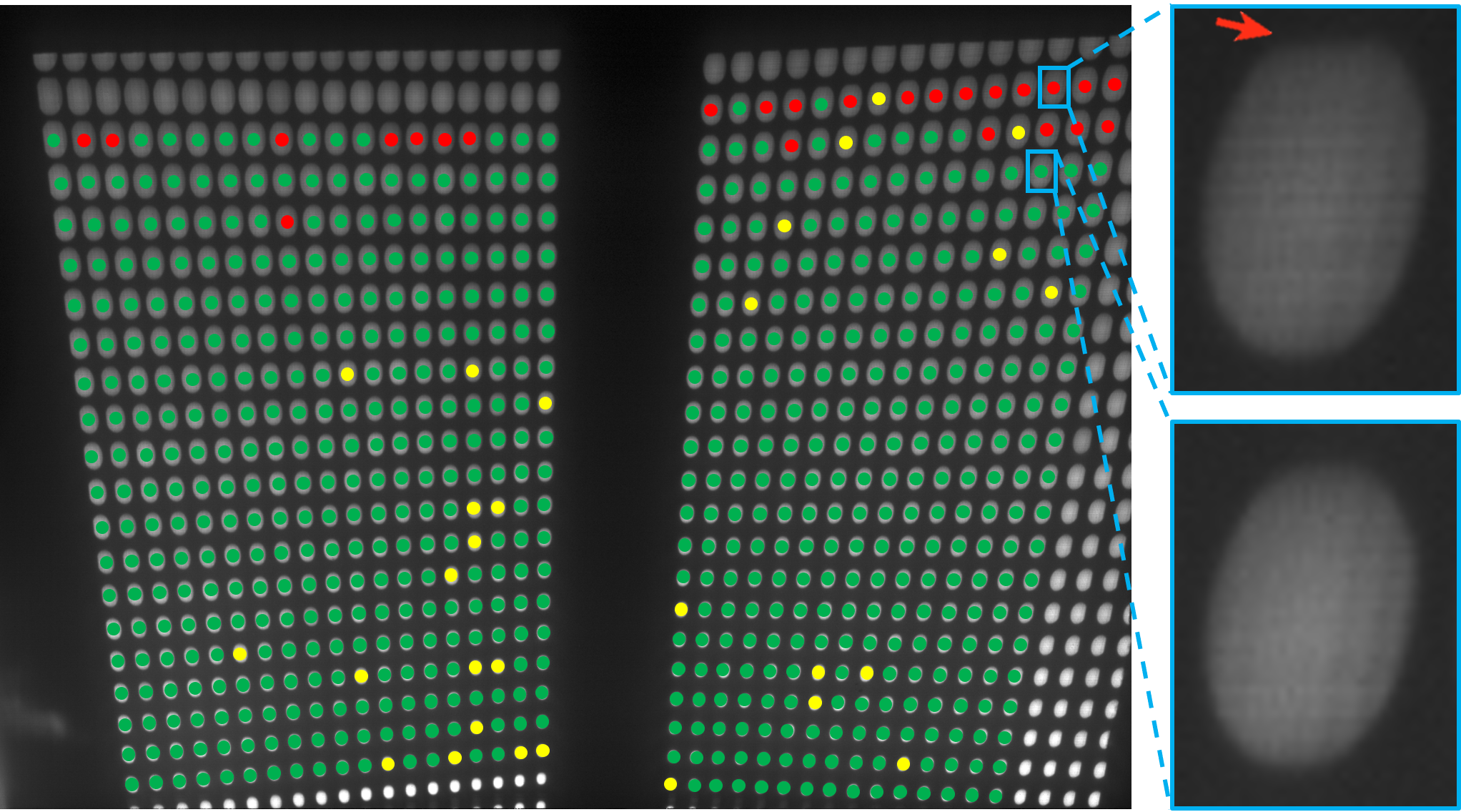}
  \caption{A scanned image when a uniform white image was projected by the LCD projector in the near condition. Green and red circles are overlaid onto inlier and outlier light blobs, respectively. Yellow circles are overlaid onto blobs whose chief ray pixels were not estimated due to missing gray code patterns. Blobs without any colored circles were excluded in the circles-grid recognition (Sect.~\ref{subsec:method_intrest}) before chief ray extraction.}
  \label{fig:outlier_analysis}
\end{figure}

Figure~\ref{fig:outlier_analysis} shows a scanned image when a uniform white image was projected by the LCD projector in the near condition.
We overlaid red circles onto the blobs that our method judged to be outliers.
In the same way, we overlaid green circles onto inlier blobs and yellow circles onto blobs whose chief ray pixels were not extracted due to missing gray codes.
From this result, we confirm that the outliers are distributed at blobs in the periphery of each pinhole-array.
The figure also shows that the outlier blobs do not form an ellipse.
This possibly occurred because the thickness of our pinhole-array masks is not infinitesimal, and, consequently, the light ray incident on a pinhole from a small grazing angle did not pass through the pinhole.
Despite the presence of outliers, intrinsic calibration was as accurate in our method as in the conventional camera-based method, as demonstrated in Sect.~\ref{subsec:exp_intrcalib}.
Therefore, we consider that the outliers do not significantly affect calibration accuracy once they are appropriately excluded.

\subsection{Dynamic projection mapping}


We conducted a dynamic PM experiment to experimentally validate the usefulness of calibration parameters estimated by our method.
The experimental setup is depicted in Fig.~\ref{fig:teaser}b.
We used the DLP1 projector in this experiment.
A 3D-printed Stanford bunny (90$\times$140$\times$130 mm) was used as a projection surface.
The bunny was tracked by an off-the-shelf motion capture system consisting of 8 cameras (NaturalPoint, OptiTrack Prime 17W).
The projector was placed approximately 3 m from the working volume.
The size of the projected image became 2.7$\times$1.5 m at a place where the bunny was moved.
We estimated the pose of the projector in the world coordinate system using the intrinsic parameters and four correspondences between the world coordinate and the projector's image coordinate, which were obtained manually.

Figure~\ref{fig:teaser}b shows dynamic PM results.
We confirmed that a projection image was rendered in each frame such that a projected texture appeared attached to the surface.
This means that both the intrinsic parameters calibrated by our method as well as the pose of the projector estimated using the intrinsic parameters were accurate enough for a dynamic PM application.
Therefore, we confirmed that our technique allows a user of a dynamic PM application to calibrate their projectors without placing a large fiducial object at different places with different poses in the large working volume of the application.

\section{Discussion}

The experimental results show that our projector calibration technique can work for different aperture sizes and focusing distances with the same calibration device of a limited dimension (320$\times$600$\times$320 mm).
Particularly, our technique could calibrate the LCD projector, while a conventional camera-based technique failed to do so because the projector and the fiducial object were too heavy and large to move by hand.
The intrinsic parameters calibrated by our system were quantitatively comparable to those calibrated using the conventional method.
We also confirmed that the parameters could be accurate enough for a dynamic PM application.

We considered the following two recommendations regarding the hardware setup to enable a potential user of our system to accurately calibrate a projector.
We found that projectors could be accurately calibrated in the condition in which all of the light blobs of projected spatial light patterns were incident on the scanner surface.
For example, when calibrating a projector with a wider FOV than those used in the experiments, it is recommended to move the projector, the pinhole-array masks, and/or the scanner so that the distances between them become shorter.
Another recommendation to a potential user of our method is to arrange the system components so that the light blobs do not overlap each other on the scanner surface.
Overlaps make it difficult to accurately identify the chief ray pixel of each light blob in the current algorithm, thus preventing a feasible calibration of the intrinsic parameters.
\revise{It is preferable that our system occupies a small space; thus, the projector should be placed as close as possible to the calibration device. However, in doing so, the light blobs can overlap. The overlap also depends on the inter-pinhole distance of the pinhole-array mask. Building a computational model to determine the most favorable projector-mask distance and inter-pinhole distance will be an important objective of future research.}

\revise{Previous camera calibration techniques that estimate the circle centers of a captured circle pattern~\cite{5457474,6466865} can be used in the chief ray extraction of our system. 
Previous techniques estimate a circle center by iteratively undistorting and unprojecting the captured image to a canonical fronto-parallel image. However, there is no theoretical guarantee that the estimation is the true circle center. On the other hand, our technique can estimate a theoretically-correct chief ray pixel. A future study could calibrate a projector by estimating the chief ray pixel using the previous technique and compare the calibration result with that obtained using the proposed technique.
Chief ray extraction becomes easier if a projector's aperture is controllable. When the lens aperture is minimized, the light blobs become single dots that are the intersections of chief rays passing through the pinholes with the scanner surface. Therefore, we do not need our back-projection technique to extract the chief rays. However, we have not found any off-the-shelf projectors whose aperture is controllable by an end user.}

Our calibration technique is limited in that it does not work for projectors whose optical systems cannot be considered pinhole projectors, such as an ultra-short throw projector that applies aspherical optics.
Conventional camera-based calibration techniques share the same limitation because they also assume a pinhole model.
Another limitation of our current prototype is that it takes a lot of time for it to calibrate a single projector.
This is primarily because it relies on a flat-bed scanner to capture the light blobs.
This problem can be solved by replacing the scanner with a flat rear-projection screen and a camera, placing the screen at the same position as the scanner surface, and capturing the appearance on the screen using the camera.
Although this setup could enable faster capturing of the light blobs than the current prototype, correspondences between physical locations on the screen and camera pixels must be precisely calibrated prior to the projector calibration.

\revise{One might argue that the intrinsic calibration could be done in the factory or lab before running an application; thus, any cost reduction achieved by the proposed technique would fail to significantly improve the workflow of dynamic PM applications. However, the focus or zoom setting of such a pre-calibrated projector would very likely change during delivery and installation. We believe that our practical calibration technique is also useful in such cases.}

\section{Conclusion}

We aimed to estimate the intrinsic parameters of a projector while avoiding the limitation of shallow DOF.
Our calibration device requires a minimal working volume directly in front of the projector lens regardless of the aperture size and focusing distance.
Our method directionally decomposes structured light using pinhole-array masks, which is then measured by a flat-bed scanner. 
We explained how to extract the chief ray pixel passing through the optical center of the projector and each pinhole.
This allows us to regard the projector as a pinhole projector and to calibrate it by applying the standard camera calibration technique that assumes a pinhole camera model.
Using a proof-of-concept prototype, we demonstrated that our technique could calibrate three projectors with different focusing distances and aperture sizes as accurately as a conventional method.
We also confirmed that our technique could provide sufficiently accurate intrinsic parameters for a large-scale dynamic PM application.
In future work, we will replace the flat-bed scanner with a camera and investigate how we can speed up calibration without sacrificing accuracy.

\acknowledgments{
The authors would like to thank Anselm Grundh\"{o}fer for valuable discussions.
This work was supported by JST, PRESTO Grant Number JPMJPR19J2, Japan.}

\bibliographystyle{abbrv-doi}

\bibliography{reference}

\appendix

\section{Conventional Projector Calibration}\label{appendix}

Our work is based on the standard camera calibration technique introduced by Zhang et al.~\cite{888718} using a pinhole camera model.
The camera and projector share the same perspective projection model, while the light directions differ.
This optical duality enables a projector to be calibrated using Zhang's technique, assuming a pinhole {\it projector} model.

Suppose $(x, y)$ is the coordinate of a pixel on the 2D image plane of a projector such as a digital micromirror device (DMD) or LCD panel, and $(X, Y, Z)$ is the coordinate of a point in 3D space of a world coordinate system, through which the projected pixel passes.
The relationship between these two coordinates is represented as follows:
\begin{equation}
    \left[
        \begin{array}{c}
            x \\ y \\ 1
        \end{array}
    \right]\propto\bm{K}[\bm{R}|\bm{t}]
    \left[
        \begin{array}{c}
            X \\ Y \\ Z \\ 1
        \end{array}
    \right],
\end{equation}
where $\bm{R}$ and $\bm{t}$ are a $3\times3$ rotation matrix and a $3\times1$ translation vector of the projector relative to the world coordinate system, respectively.
$\bm{K}$ is the projector's intrinsic matrix and
is defined as follows:
\begin{equation}
    \bm{K}=\left[
        \begin{array}{ccc}
            f_x & 0 & c_x \\
            0 & f_y & c_y \\
            0 & 0 & 1
        \end{array}
    \right],   
\end{equation}
where $f_x$ and $f_y$ are, respectively, the horizontal and vertical focal lengths of the projector.
$c_x$ and $c_y$ are the coordinates of the projector’s principal point on the image plane.
Most camera calibration algorithms also take into account the radial and tangential lens distortions.
On the other hand, thanks to advances in manufacturing, such non-linear distortions in projectors are generally unnoticeable.
Therefore, as a reasonable assumption for the vast majority of existing modern projectors, we do not consider distortions and focus instead only on calibrating the parameters of the intrinsic matrix, though our method can also deal with them.

%
%

A general procedure of projector calibration, extended from Zhang's method, is as follows.
First, a projector to be calibrated projects a spatial pattern or a series of structured light patterns onto a flat surface on which a 3D coordinate system is defined.
Second, we obtain correspondences between the 3D coordinate of each point on the surface (object point) and the 2D coordinate of a projected pixel (image point).
The correspondences are obtained on multiple surfaces of different poses.
Then, the projector's parameters are initially estimated by solving the closed-form solution using the homographies between the flat surfaces and the projector's image plane.
This result is further refined in a non-linear optimization step where an optimization method such as the Levenberg-Marquardt algorithm minimizes the reprojection error (i.e., the sum of the squared distances between the image points and the projected corresponding object points that are mathematically projected to the projector's image plane using the estimates at each iteration of the optimization).

\end{document}